\documentclass{article} 
\usepackage{iclr2021_conference,times}

\usepackage{amsmath,amsfonts,bm}

\def\eqref#1{equation~\ref{#1}}

\def\1{\bm{1}}

\DeclareMathAlphabet{\mathsfit}{\encodingdefault}{\sfdefault}{m}{sl}
\SetMathAlphabet{\mathsfit}{bold}{\encodingdefault}{\sfdefault}{bx}{n}

\usepackage{hyperref}
\usepackage{url}
\usepackage[pdftex]{graphicx}

\title{Learning to Act: Novel Integration of Algorithms and Models for Epidemic Preparedness
}

\author{Sekou L. Remy \& Oliver E. Bent \\
IBM Research Africa\\
Nairobi, Kenya \\
sekou@ke.ibm.com, oliver.bent2@ibm.com \\
}

\usepackage{listings}
\usepackage[normalem]{ulem}
\usepackage{algorithm2e} 
\usepackage{xcolor}
\definecolor{dkgreen}{rgb}{0,0.6,0}
\definecolor{gray}{rgb}{0.5,0.5,0.5}
\definecolor{mauve}{rgb}{0.58,0,0.82}

 \usepackage[caption=false,font=normalsize,labelfont=sf,textfont=sf]{subfig}

\iclrfinalcopy 
\begin{document}

\maketitle

\begin{abstract}In this work we present a framework which may transform research and praxis in epidemic planning.
Introduced in the context of the ongoing COVID-19 pandemic, we provide a concrete demonstration of the way algorithms may learn from epidemiological models to scale their value for epidemic preparedness. 
Our contributions in this work are two fold: 1) a novel platform which makes it easy for decision making stakeholders to interact with epidemiological models and algorithms developed within the Machine learning community, and 2) the release of this work under the Apache-2.0 License.
The objective of this paper is not to look closely at any particular models or algorithms, but instead to highlight how they can be coupled and shared to empower evidence-based decision making.

\end{abstract}

\section{Introduction}
This work provides a concrete example of a new paradigm to inform decision support processes in a public health context.
Public health professionals are expected to engage in evidence-informed decision making to advise their practice \citep{petticrew2004evidence, yost2014tools}.
As indicated in \citet{nutbeam2001evidence}, evidence is required from a variety of sources including expert knowledge, existing domestic and international research, stakeholder consultation, and even assessment of existing policies.
While it is rational to expect evidence to be used, challenges in evidence based decision-making are ever present. 
Today, very few decisions have complete information, the simplest decisions often have innumerable outcomes which are ultimately uncertain. 
For high impact decision making, such as epidemic planning, though of great importance, it is not possible to evaluate all possible options, or completely characterize their uncertainty.
Many turn to modeling as a mechanism to generate insights about what could potentially happen, and consider model output as informative in the decision making process.
It is well known that all models can be considered to be `wrong' \citep{ioannidis2020forecasting}, and in some cases have led to ill conceived actions being considered, but it has also been shown that when properly contextualized, even these incorrect models can be useful \citep{holmdahl2020wrong}.

In this work we demonstrate an approach to harness epidemiological models to generate evidence which can inform decision making.
We utilize Machine learning algorithms from a range of classes (from RL, to Optimization, to Planning).
Our aim is not to focus on model validity as defined in \citet{petticrew2004evidence}, nor the non-technical issues related to adoption of the insights generated from these models \citet{liverani2013political}, nor even to adjudicate which model (or algorithm) is `best'.
Our goal is to show the community how models can be connected with these algorithms in a flexible manner which will support integration of multiple classes and implementations of models.
What we champion here will also provide a mechanism to realize a repeatable infrastructure to generate insights, not just for COVID-19, but for any decision making process which can be informed by multiple models and data sources. 
The resources we describe in this work have been released via a public GitHub source repository\footnote{https://github.com/IBM/ushiriki-policy-engine-library} which serves as both an installation point and templates for continued extension.

\section{Methodology}
In this project we utilize the framework presented in \cite{notnets} to access models at scale with transparency and trust.
This framework enables API mediated access to a suite of models.
Containerized models are deployed on-demand, and configured at runtime to use the input parameters defined by the user.
The framework marshals the use of the model and the associated data in a manner that gives both the users and owners of these assets confidence in the output which is generated.
Moreover, the framework permits data collected by different algorithms, and possibly even at different times, to be shared. 
Every additional execution of an algorithm contributes to the pool of knowledge which exists about the models.
Finally, the framework also supports the flexible integration of input into a format which can be consumed by the specific models which are available.

For the larger project in which this work is situated, we select models provided by modelers from different communities. 
In this manuscript we will only highlight the use of two types of models:  compartmental models \citep{brauer2008compartmental, edlund2011comparing} and agent based models \citep{macal2009agent, Kerr2020.05.10.20097469} as they are both widely used by epidemiologists across a variety of domains.
For these instances of COVID-19 models referenced in this work, the countermeasures (or interventions) are captured as sequences of parameter values.
In the first framing, we permit the actions (e.g. physical distancing, mask wearing, vaccination) to directly change model parameters like transmission rate, recovery rate, and the death rate.
In an alternate framing, we also link model parameters to the Stringency Index \citep{hale2020variation}, a metric which captures the impact that a particular set of countermeasures has on disease progression. 
These framings represent distinct implementations of models specifically either for model predictions or assessing the impact of control policies.

\begin{figure}
\centering
\includegraphics[width=.4\textwidth]{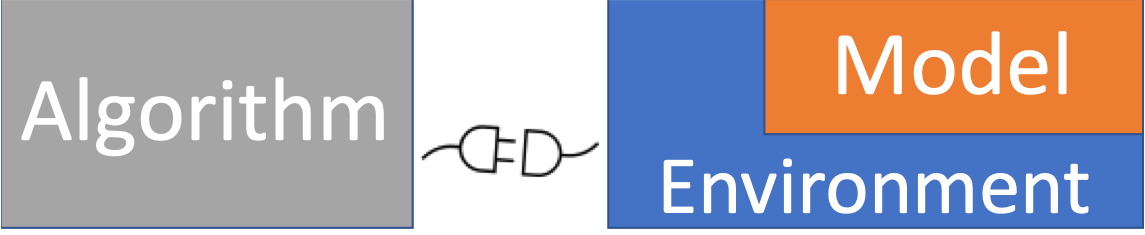}
\caption{The key abstraction presented in this work wraps domain models in OpenAI Gym environments, exposing the STEP method as a connection between models and algorithms.
}
\label{fig:abstraction}
\end{figure}
\vspace{-.5cm}
\begin{lstlisting}[caption={Sample code instantiating an environment with access of the STEP method on line \#4.}, float={h}, label=sample]
import gym 
env = gym.make(name, userID=userID, baseuri=baseuri)
env.reset()
observation, reward, done, _ =  env.step(env.action_space.sample())
\end{lstlisting}
\vspace{-.5cm}
The crux of the contribution of this work lies in encapsulating the epidemiological model into an OpenAI Gym environment.
In this case, when the step method is provided with an action (see Listing \ref{sample}), the model is able to run and provides information indicating what occurred (the observation), and a metric capturing the goodness of the action (the reward).
Other metadata is provided as well, but it is out of the scope of this paper.
Depending on the definition of the environment, the action can be a single intervention or a sequence of interventions.

In this project we have successfully harnessed algorithms from multiple domains of the Machine Learning spectrum. 
From Reinforcement Learning \citep{watkins1992q, keerthi1994tutorial, mnih2013playing} to Optimization \citep{Srinivas2009, Contal2013}, they have all been implemented utilizing the \textit{step} method of the environment to access the reward function (and in some cases the observations of the model states) when presented with an action/intervention to evaluate.
An example of these algorithms in seen in the psuedocode for Bayesian optimization presented in Algorithm \ref{bayes_opt}.
Due to space constraints, we omit discussion of the algorithm itself, but refer the curious reader to \citet{Contal2013} for further details.
\begin{algorithm}
\label{bayes_opt}
\caption{An Implementation of Bayesian Optimization using the STEP method.}
\KwResult{Policy $\pi_{greedy} = \mbox{argmax}\left(\hat{v}_{\pi}(\theta^{gp})\right)$, where $\hat{v}_{\pi}(\theta^{gp}) \approx v_\pi$ }
Initialise $\mathcal{GP}$ priors, $\theta^{GP}_0$
 \For{i = 1,2, ..., end}{
 \eIf{i $\leq$ k}{select $\pi_i = rand(\pi)$}
 {select $\pi_i = \mbox{argmax}(\alpha(\hat{v}_{\pi}(\theta^{gp}_{i-1})))$, where $\alpha$ is the acquisition function}
 $R_i$ = \textit{step}($\pi_{i}$)
 }
Update $\mathcal{GP}$ Posterior $\hat{v}(\theta^{gp}_i)$; mean $m_{i}(\pi)$; co-variance function $k_{i}(\pi)$;
\end{algorithm}

\section{Results}
\textit{What are the most cost effective ways to ``flatten the curve''?
If there is a fixed amount of money which can be allocated to interventions over the next three months, what sequence would be best?
If you wish to simplify your policies, which reduced sets of interventions are most impactful?}
To address any of these questions using model derived evidence, one must first calibrate the model to data for the location in question, and then one can utilize an environment which provides the reward functions relevant to the particular question.

\begin{figure*}[ht]
\centering
\subfloat[font=footnotesize][Agent Based Model]{
\includegraphics[width=.32\textwidth]{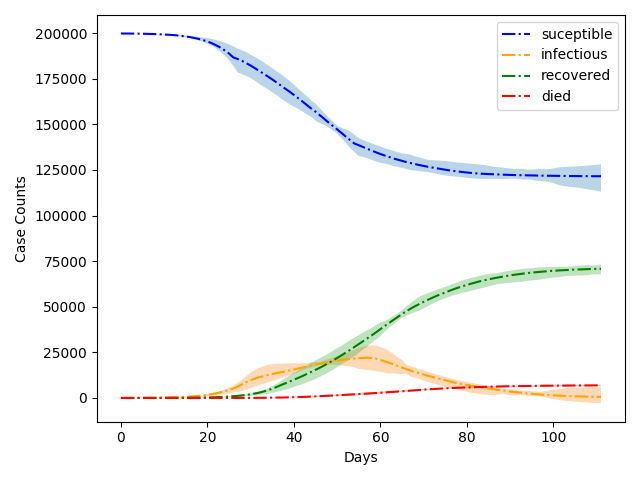}
\label{fig:covasimcalib}
}\hfill
\subfloat[font=footnotesize][Compartmental Model with Stringency]{
\includegraphics[width=.32\textwidth]{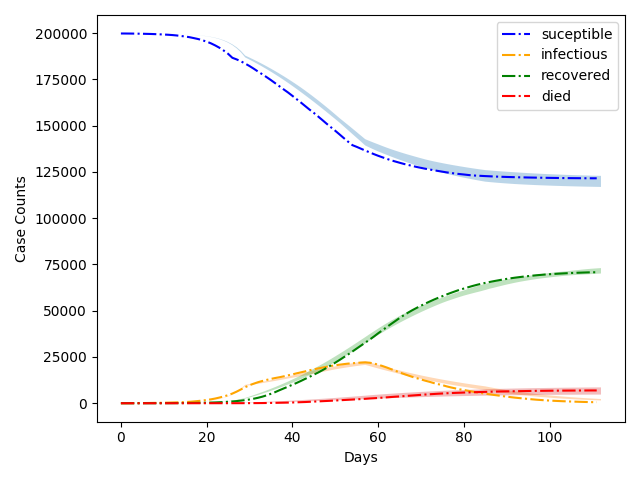}
\label{fig:sirdcalib}
}\hfill
\subfloat[font=footnotesize][Compartmental Model]{
\includegraphics[width=.32\textwidth]{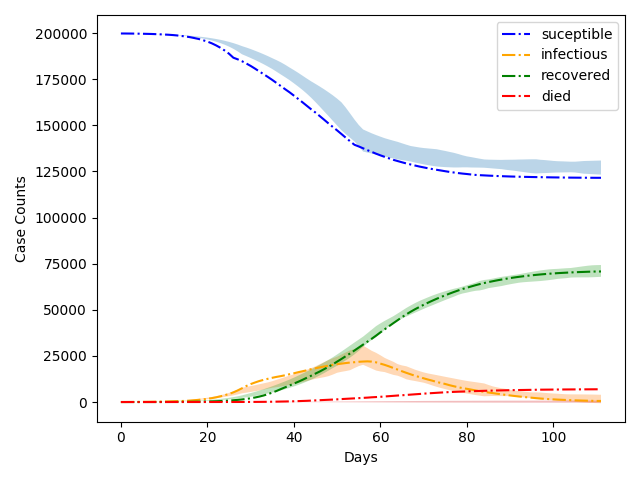}
\label{fig:sirdsicalib}
}
\caption{Calibration of three environments to the same case data (in this case using more data than is typically available).
In each calibration, the results consider learning 20 sets of parameters.
}
\label{fig:calibrations}
\end{figure*}

As indicated previously, the same algorithms can be applied with the multiple algorithm types and model types (through the use of the respective environments).
Moreover, for each model, there were environments for calibration and optimization with various objective functions.
A subset of the combinations is included herein for brevity.
Figure \ref{fig:calibrations} shows the results of calibrating the parameters three different models to the same data, utilizing Algorithm \ref{bayes_opt}.
In each case the algorithm was able to identify reasonable fits, even though the structures of the models were quite different.

\begin{figure}[ht]
\centering
\subfloat[][Lenient]{
\includegraphics[width=.4\columnwidth, height = 5cm]{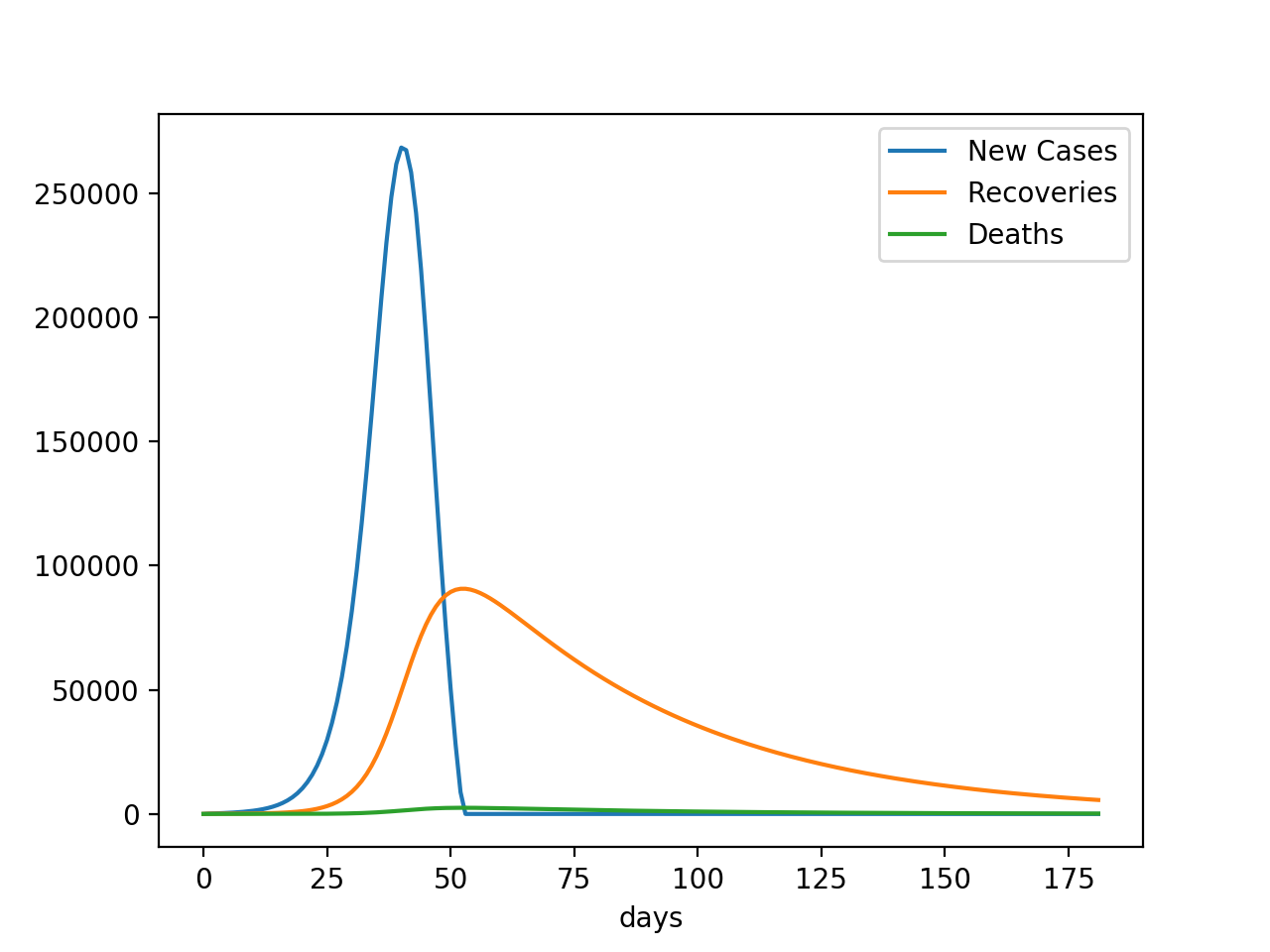}
\label{fig:90}
}\quad
\subfloat[][Stringent]{
\includegraphics[width=.4\columnwidth, height = 5cm]{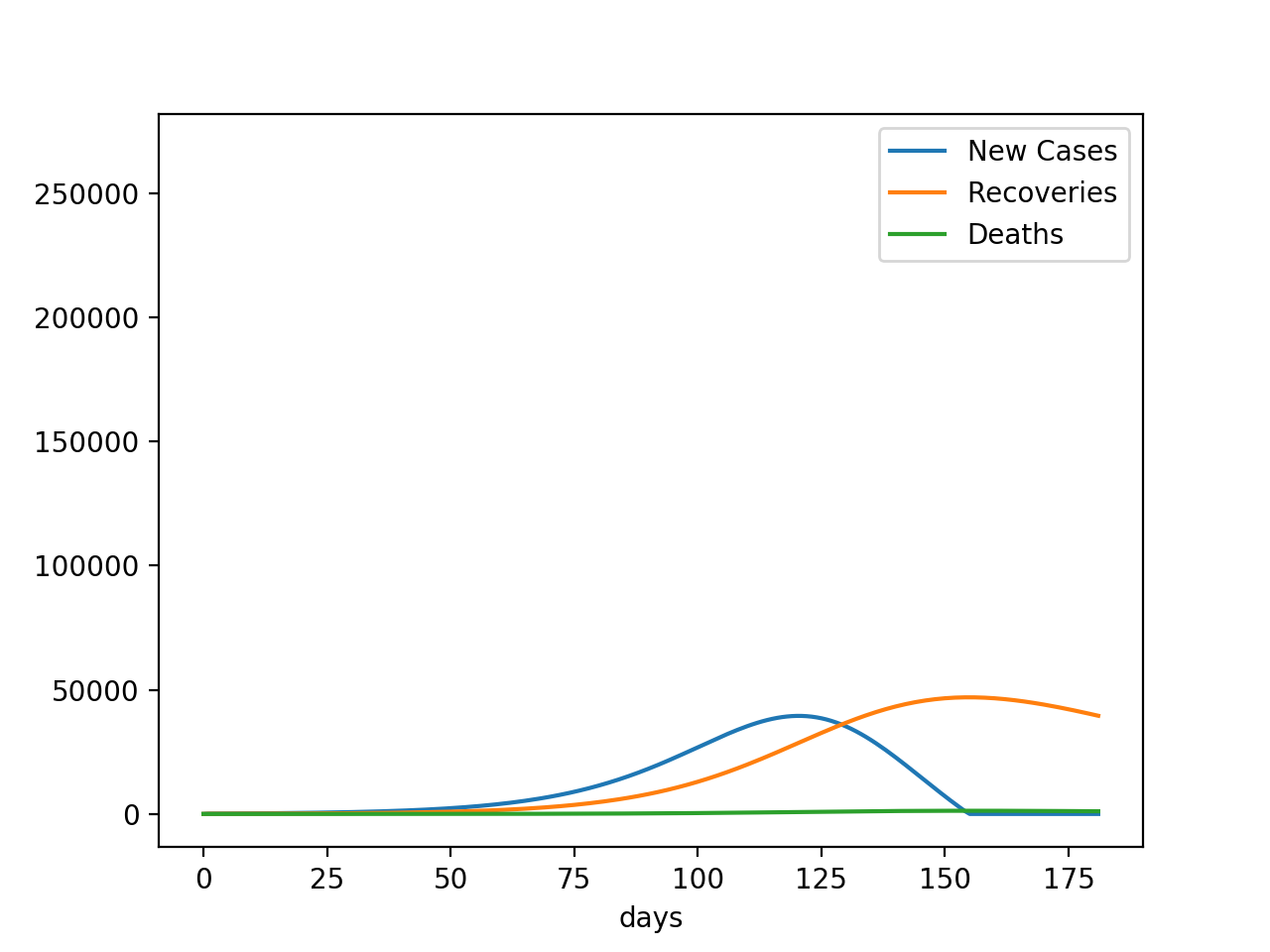}
\label{fig:09}
}
\caption{Two examples of deterministic policies of a single countermeasure for the simulated timeframe.
Could a reactive policy have performed ``better'' according to a chosen reward function?
}
\label{fig:policies}
\end{figure}

The reward function in the case of calibration is the error between the data and the observed model output.
The actions permit the algorithms to set the values of the calibration parameters.
By changing the reward function to represent different properties, for example, the number of cases, and changing the actions to modify the transmission rate, another environment can be created permitting the same models to help consider the problem of flattening the curve.
In this case, the algorithm's role is to find the best set of interventions (the sequence of transmission rates) which minimize the number of individuals who contract the disease (see Figure \ref{fig:policies}).

In this case, the environment's states are the values from each of the compartments (or the fraction of the population in a given state); 
the rewards produced are the cumulative incidence for the two week period which each step is assessed for; 
and the actions are integers in the range $[0,99]$. 
The overall goal is thus to minimize the total number of cases by learning when to vary the stringency of the interventions implemented.
In Figure \ref{fig:policies}, we show the state trajectories (incidence) for two sequences of actions.
The first was a very stringent disease management response which was learned, while the other may be considered lenient.
In both cases there are individuals in the population who contract the disease, however under the stringent policy, there are much less who do.
Also, under this policy, the peak number who are infected at the same time is also lower.
This might be considered as an example of the impact of ``flattening the curve''.
Using this type of approach, and by changing the reward function, it would be possible to raise the question of whether there are other intervention programs (sequences of actions) which could have flattened the curve with a lower societal cost.

As alluded, by changing the reward function but keeping all the actions the same (which changes the response from the STEP function) we can also consider the impact of sequences of actions under a different rubric.
In this case we quantify the impact of actions on the community.
Executing interventions have a cost, but there's also a cost if individuals in the populations get infected.
Individuals can be infected even when the most stringent interventions are put into place.
Health economists have characterized the notion of Years of effective Life Lost as one such measure to quantify the impact of an illness.
In this work we implement what they have outlined and also include the cost of an intervention inferred from the newspaper sources.

\section{Summary}

In proposing a new paradigm to inform decision support for epidemic planning, we have detailed extensions of epidemiological models to learning environments and the demonstration of learning algorithm performance on such environments. 
We posit that there is much to consider in the rich space of reward functions which arise from capturing the complexity of decisions in the real-world which are full of trade-offs and balancing competing needs.
It is left to be seen whether the algorithms which will be most successful in this problem domain will be similar to those which are dominant in domains such as backgammon, chess, or even those which have physics based dynamics like the cart-pole or walking robots. 
COVID-19 has shown us that there is still need for research advancement in the process of decision-making being supported by models.

Our primary contribution in this work is the demonstration of how decision makers can access self-contained resources curated by the relevant experts (e.g. epidemiological modelers and ML experts). 
The secondary, and equally as valuable contribution is the presentation of the open sourced, OpenAI Gym environments typifying the encapsulation of epidemiological models in a manner which will be illustrative and enable stakeholders to play with the assets themselves, and subsequently to extend them to include models/data/algorithms which are relevant for their context.

In closing, we acknowledge the limitations of models but we still advocate for their use in principled and pragmatic ways.
We specifically assert that whether considered in the traditional sense or as we've proposed in this work, any model usage must be in line with the intended manner defined by the model creators, as well as those defined by the users who will consume the output of the model.
While we connect models to algorithms by extending the model abstractions to environments, we take care not to violate the underlying assumptions and definitions prescribed by their creators.
In our future work we will perform user studies with the aim of characterizing the utility of these resources in practice, and to better understand how to amplify the decision making power of policy makers.
Further, we will use in silico means to consider the impact of data integrity on the calibration and optimization results which can be generated using our approach. 

\newpage
\bibliography{iclr2021_conference}

\begin{thebibliography}{17}
\providecommand{\natexlab}[1]{#1}
\providecommand{\url}[1]{\texttt{#1}}
\expandafter\ifx\csname urlstyle\endcsname\relax
  \providecommand{\doi}[1]{doi: #1}\else
  \providecommand{\doi}{doi: \begingroup \urlstyle{rm}\Url}\fi

\bibitem[Brauer(2008)]{brauer2008compartmental}
Fred Brauer.
\newblock Compartmental models in epidemiology.
\newblock In \emph{Mathematical epidemiology}, pp.\  19--79. Springer, 2008.

\bibitem[Contal et~al.(2013)Contal, Buffoni, Robicquet, and
  Vayatis]{Contal2013}
Emile Contal, David Buffoni, Alexandre Robicquet, and Nicolas Vayatis.
\newblock {Parallel Gaussian process optimization with upper confidence bound
  and pure exploration}.
\newblock \emph{Lecture Notes in Computer Science (including subseries Lecture
  Notes in Artificial Intelligence and Lecture Notes in Bioinformatics)}, 8188
  LNAI\penalty0 (PART 1):\penalty0 225--240, 2013.
\newblock ISSN 03029743.
\newblock \doi{10.1007/978-3-642-40988-2_15}.

\bibitem[Edlund et~al.(2011)Edlund, Kaufman, Lessler, Douglas, Bromberg,
  Kaufman, Bassal, Chodick, Marom, Shalev, et~al.]{edlund2011comparing}
Stefan Edlund, James Kaufman, Justin Lessler, Judith Douglas, Michal Bromberg,
  Zalman Kaufman, Ravit Bassal, Gabriel Chodick, Rachel Marom, Varda Shalev,
  et~al.
\newblock Comparing three basic models for seasonal influenza.
\newblock \emph{Epidemics}, 3\penalty0 (3-4):\penalty0 135--142, 2011.

\bibitem[Hale et~al.(2020)Hale, Petherick, Phillips, and
  Webster]{hale2020variation}
Thomas Hale, Anna Petherick, Toby Phillips, and Samuel Webster.
\newblock Variation in government responses to covid-19.
\newblock \emph{Blavatnik School of Government Working Paper}, 31, 2020.

\bibitem[Holmdahl \& Buckee(2020)Holmdahl and Buckee]{holmdahl2020wrong}
Inga Holmdahl and Caroline Buckee.
\newblock Wrong but useful—what covid-19 epidemiologic models can and cannot
  tell us.
\newblock \emph{New England Journal of Medicine}, 2020.

\bibitem[Ioannidis et~al.(2020)Ioannidis, Cripps, and
  Tanner]{ioannidis2020forecasting}
John~PA Ioannidis, Sally Cripps, and Martin~A Tanner.
\newblock Forecasting for covid-19 has failed.
\newblock \emph{International journal of forecasting}, 2020.

\bibitem[Keerthi \& Ravindran(1994)Keerthi and Ravindran]{keerthi1994tutorial}
S~Sathiya Keerthi and B~Ravindran.
\newblock A tutorial survey of reinforcement learning.
\newblock \emph{Sadhana}, 19\penalty0 (6):\penalty0 851--889, 1994.

\bibitem[Kerr et~al.(2020)Kerr, Stuart, Mistry, Abeysuriya, Hart, Rosenfeld,
  Selvaraj, Nunez, Hagedorn, George, Izzo, Palmer, Delport, Bennette, Wagner,
  Chang, Cohen, Panovska-Griffiths, Jastrzebski, Oron, Wenger, Famulare, and
  Klein]{Kerr2020.05.10.20097469}
Cliff~C. Kerr, Robyn~M. Stuart, Dina Mistry, Romesh~G. Abeysuriya, Gregory
  Hart, Katherine Rosenfeld, Prashanth Selvaraj, Rafael~C. Nunez, Brittany
  Hagedorn, Lauren George, Amanda Izzo, Anna Palmer, Dominic Delport, Carrie
  Bennette, Bradley Wagner, Stewart Chang, Jamie~A. Cohen, Jasmina
  Panovska-Griffiths, Michal Jastrzebski, Assaf~P. Oron, Edward Wenger, Michael
  Famulare, and Daniel~J. Klein.
\newblock Covasim: an agent-based model of covid-19 dynamics and interventions.
\newblock \emph{medRxiv}, 2020.
\newblock \doi{10.1101/2020.05.10.20097469}.
\newblock URL
  \url{https://www.medrxiv.org/content/early/2020/05/15/2020.05.10.20097469}.

\bibitem[Liverani et~al.(2013)Liverani, Hawkins, and
  Parkhurst]{liverani2013political}
Marco Liverani, Benjamin Hawkins, and Justin~O Parkhurst.
\newblock Political and institutional influences on the use of evidence in
  public health policy. a systematic review.
\newblock \emph{PloS one}, 8\penalty0 (10):\penalty0 e77404, 2013.

\bibitem[Macal \& North(2009)Macal and North]{macal2009agent}
Charles~M Macal and Michael~J North.
\newblock Agent-based modeling and simulation.
\newblock In \emph{Proceedings of the 2009 Winter Simulation Conference (WSC)},
  pp.\  86--98. IEEE, 2009.

\bibitem[Mnih et~al.(2013)Mnih, Kavukcuoglu, Silver, Graves, Antonoglou,
  Wierstra, and Riedmiller]{mnih2013playing}
Volodymyr Mnih, Koray Kavukcuoglu, David Silver, Alex Graves, Ioannis
  Antonoglou, Daan Wierstra, and Martin Riedmiller.
\newblock Playing atari with deep reinforcement learning.
\newblock \emph{arXiv preprint arXiv:1312.5602}, 2013.

\bibitem[Nutbeam(2001)]{nutbeam2001evidence}
Don Nutbeam.
\newblock Evidence-based public policy for health: matching research to policy
  need.
\newblock \emph{Global Health Promotion}, pp.\ ~15, 2001.

\bibitem[Petticrew et~al.(2004)Petticrew, Whitehead, Macintyre, Graham, and
  Egan]{petticrew2004evidence}
Mark Petticrew, Margaret Whitehead, Sally~J Macintyre, Hilary Graham, and Matt
  Egan.
\newblock Evidence for public health policy on inequalities: 1: the reality
  according to policymakers.
\newblock \emph{Journal of Epidemiology \& Community Health}, 58\penalty0
  (10):\penalty0 811--816, 2004.

\bibitem[Srinivas et~al.(2009)Srinivas, Krause, Kakade, and
  Seeger]{Srinivas2009}
Niranjan Srinivas, Andreas Krause, Sham~M Kakade, and Matthias Seeger.
\newblock {Gaussian Process Optimization in the Bandit Setting: No Regret and
  Experimental Design}.
\newblock 2009.
\newblock ISSN 00189448.
\newblock \doi{10.1109/TIT.2011.2182033}.
\newblock URL
  \url{http://arxiv.org/abs/0912.3995{\%}7B{\%}25{\%}7D0Ahttp://dx.doi.org/10.1109/TIT.2011.2182033}.

\bibitem[Wachira et~al.(2020)Wachira, Remy, Bent, Bore, Osebe, Weldemariam, and
  Walcott-Bryant]{notnets}
Charles Wachira, Sekou~L Remy, Oliver Bent, Nelson Bore, Samuel Osebe,
  Komminist Weldemariam, and Aisha Walcott-Bryant.
\newblock A platform for disease intervention planning.
\newblock In \emph{2020 IEEE International Conference on Healthcare Informatics
  (ICHI) - to appear}. IEEE, 2020.

\bibitem[Watkins \& Dayan(1992)Watkins and Dayan]{watkins1992q}
Christopher~JCH Watkins and Peter Dayan.
\newblock Q-learning.
\newblock \emph{Machine learning}, 8\penalty0 (3-4):\penalty0 279--292, 1992.

\bibitem[Yost et~al.(2014)Yost, Dobbins, Traynor, DeCorby, Workentine, and
  Greco]{yost2014tools}
Jennifer Yost, Maureen Dobbins, Robyn Traynor, Kara DeCorby, Stephanie
  Workentine, and Lori Greco.
\newblock Tools to support evidence-informed public health decision making.
\newblock \emph{BMC public health}, 14\penalty0 (1):\penalty0 728, 2014.

\end{thebibliography}
\bibliographystyle{iclr2021_conference}

\end{document}